\documentclass[runningheads]{llncs}
\usepackage[T1]{fontenc}
\usepackage{textcomp}
\usepackage{stfloats}
\usepackage{url}
\usepackage{verbatim}
\usepackage{graphicx,wrapfig,lipsum}
\usepackage{booktabs}
\usepackage{xcolor}
\usepackage{multicol}
\usepackage{times}
\usepackage{epsfig}
\usepackage{float}
\usepackage{amssymb}
\usepackage{algorithm}
\usepackage{hyperref}
\usepackage{lipsum}
\usepackage{dirtytalk}
\usepackage{algpseudocode}
\usepackage{adjustbox}
\usepackage{multirow}

\begin{document}
\title{Local Neighborhood Features for 3D Classification}
%
%\titlerunning{Abbreviated paper title}
% If the paper title is too long for the running head, you can set
% an abbreviated paper title here
%

% AUTHORS AND INSTITUTIONS COMMENTED OUT FOR ANONYMITY
% \author{Paper ID 98}
\author{Shivanand Venkanna Sheshappanavar\orcidID{0000-0003-4039-2910} \and
% Second Author\inst{2,3}\orcidID{1111-2222-3333-4444} \and
Chandra Kambhamettu\orcidID{0000-0001-5306-3994}}
%%
%\authorrunning{F. Author et al.}
%% First names are abbreviated in the running head.
%% If there are more than two authors, 'et al.' is used.
%%
% \institute{Anonymous}
\institute{Video/Image Modeling and Synthesis (VIMS) Lab, University of Delaware,\\ 212 Smith Hall, 18 Amstel Ave, Newark, DE 19711, USA\\
\email{ssheshap@udel.edu, chandrak@udel.edu}\\
\url{http://bigdatavision.org/}}
% \and
% ABC Institute, Rupert-Karls-University Heidelberg, Heidelberg, Germany\\
% \email{\{abc,lncs\}@uni-heidelberg.de}}
%
\maketitle              % typeset the header of the contribution
\begin{abstract}
With advances in deep learning training strategies, the training of Point cloud classification methods is significantly improving. For example, PointNeXt, which adopts prominent training techniques and InvResNet layers into PointNet++, achieves over 7\% improvement on the real-world ScanObjectNN dataset. A typical set abstraction layer in these models maps point coordinate features of neighborhood points to higher dimensional space and passes point coordinates to the next set abstraction layer. However, these models ignore already computed neighborhood point features as additional neighborhood features. In this paper, we revisit the PointNeXt model to study the usage and benefit of neighborhood point distance and directional vectors as additional neighborhood features. We train and evaluate PointNeXt on ModelNet40 (synthetic), ScanObjectNN (real-world), and a recent large-scale, real-world grocery dataset, i.e., 3DGrocery100. In addition, we provide an additional inference strategy of weight averaging the top two checkpoints of PointNeXt to improve classification accuracy. Together with the above-mentioned ideas, we gain \textbf{0.5\%}, \textbf{1\%}, \textbf{4.8\%}, \textbf{3.4\%}, \textbf{1.6\%}, and \textbf{2.8\%} overall accuracy on the PointNeXt model with real-world datasets, ScanObjectNN (hardest variant), 3DGrocery100's Apple10, Fruits, Vegetables, Packages, and Full subsets, respectively. We also achieve a comparable~\textbf{0.2\%} accuracy gain on ModelNet40. Finally, we provide detailed ablation studies discussing the trade-offs of using additional neighborhood features. Code is available at \url{https://github.com/VimsLab/Local3DFeatures}.

\end{abstract}
\section{Introduction}
3D Computer Vision is a vibrant research domain with broad applications in augmented/virtual reality and autonomous-driving vehicles. Over the past five years, the application of deep neural networks to 3D computer vision, especially for 3D point cloud processing, has progressed tremendously. This progress has led to state-of-the-art results on several computer vision tasks, such as 3D Object Classification, 3D Semantic Segmentation, 3D Scene Understanding, and 3D Shape retrieval. \hfill\break
\indent Point cloud Object Classification has gained traction since the pioneering work of PointNet~\cite{qi2017pointnet} that processed raw point sets through Multi-Layer Perceptrons (MLPs). However, PointNet, while aggregating features at the global level using max-pooling operation, lost valuable local geometric information. Overcoming this limitation, PointNet++~\cite{qi2017pointnet++} employed ball querying and/or k-Nearest Neighbor (k-NN) querying to query local neighborhoods to extract local semantic information. 

\indent Although PointNet++~\cite{qi2017pointnet++} was effective in capturing local geometric information, it still lost contextual information due to the max-pooling operation. Several methods~\cite{qi2017pointnet++,liu2019relation,hua2018pointwise,duan2019structural,liu2019densepoint,xu2020geometry,lan2019modeling} since PointNet++ have used raw $x,y,z$ point coordinates as input. First, sample farthest points (FPS) from the input points, and at each FPS point, query neighborhood points using a fixed radius ball, i.e., compute the distance to each point from each of the FPS/anchor points to check if the point is within the neighborhood of the FPS point. Secondly, group a small sample of neighborhood points at each FPS point based on this distance. From the queried neighborhood points, subtracting the anchor FPS point gives directional vectors of the neighborhood. Finally, these directional vectors are mapped to higher dimensions in each set abstraction layer (SA) to get local features. After this step, the directional vectors and the computed distance (in the case of ball querying) are ignored. 

The most recent model, PointNeXt~\cite{PointNeXt}, builds upon the PointNet++ model, which uses ball querying for neighborhood querying by computing the distance from the anchor point to the neighborhood point to check if the point is within a ball of radius $r$. After querying, PointNeXt computes the directional vectors from the grouped neighborhood points, normalizes the vectors using the radius, and encodes them into local neighborhood features but does not use the distance and the directional vectors as additional local features. This paper emphasizes using the radius-normalized distance and the directional vectors as additional local neighborhood features with minimal additional memory or computational costs. The contributions of our work are four-fold:

\begin{itemize}
\item We use radius $r$-normalized neighborhood point distance as an additional neighborhood feature to improve the classification accuracy.
\item We show radius $r$-normalized directional vectors as additional neighborhood features benefit several models such as PointNeXt~\cite{PointNeXt}.
\item We demonstrate that averaging the weights of two best model checkpoints (models saved in the same training session) benefits test/inference accuracy.
\item We train and evaluate PointNeXt with our combined approaches on one synthetic, i.e., ModelNet40~\cite{wu20153d}, and two real-world datasets, i.e., ScanObjectNN~\cite{uy2019revisiting} and 3DGrocery100~\cite{sheshappanavar2024benchmark}. 
\end{itemize}

\section{Related Works}

PointNet++~\cite{qi2017pointnet++} is composed of Set Abstraction (SA) Layers, and each SA constitutes three layers. A Farthest Point Sampling (FPS) layer to sample a uniformly distributed subset of points in the input point cloud. A grouping layer partitions the input point cloud into common hierarchical structures. The third layer, a pointnet layer, learns contextual representation from the grouped points.

Using Kernel Correlation, KCNet~\cite{shen2018mining} mines local points and measures point affinity with similar geometric structures. Different convolution operations~\cite{bronstein2017geometric} capture different levels of geometric information from the surface deformations. Using the EdgeConv technique, Dynamic Graph Convolutional Neural Network (DGCNN)~\cite{wang2019dynamic} captures local geometric features (kNN is used for neighborhood querying). PointGrid~\cite{le2018pointgrid} proposed an integrated point and grid hybrid 3D CNN model to represent the local geometry better. DensePoint~\cite{liu2019densepoint} recursively concatenates features from MLPs to learn sufficient contextual semantic information. 

Relation Shape Convolutional Neural Network (RS-CNN)~\cite{liu2019relation} proposed a relation-shape convolution to explicitly encode the geometric relation of points for better awareness of the underlying shape. Although RS-CNN uses the Euclidean distance as one of the relations, it does not normalize the distance with the ball radius. The critical difference between our approach and RS-CNN is that our approach normalizes the distance (between the neighborhood point and its anchor point) with the radius $r$.

Although novel neighborhood querying methods~\cite{sheshappanavar2020novel,sheshappanavar2021dynamic} have demonstrated the use of Eigenvalues of the neighborhood as additional features, they devise a two-pass querying of the neighborhood, first with a ball and then using an oriented (and scaled) ellipsoid resulting in extra training and inference time. While PointNeXt only normalizes directional vectors, it does not include the vectors as additional features after mapping them to higher dimensions. Our approach uses these normalized directional vectors as neighborhood features along with the radius $r$-normalized distance.

\begin{figure}[htbp]
\begin{center}
%\fbox{\rule{0pt}{2in} \rule{0.9\linewidth}{0pt}}
\includegraphics[width=0.9\linewidth]{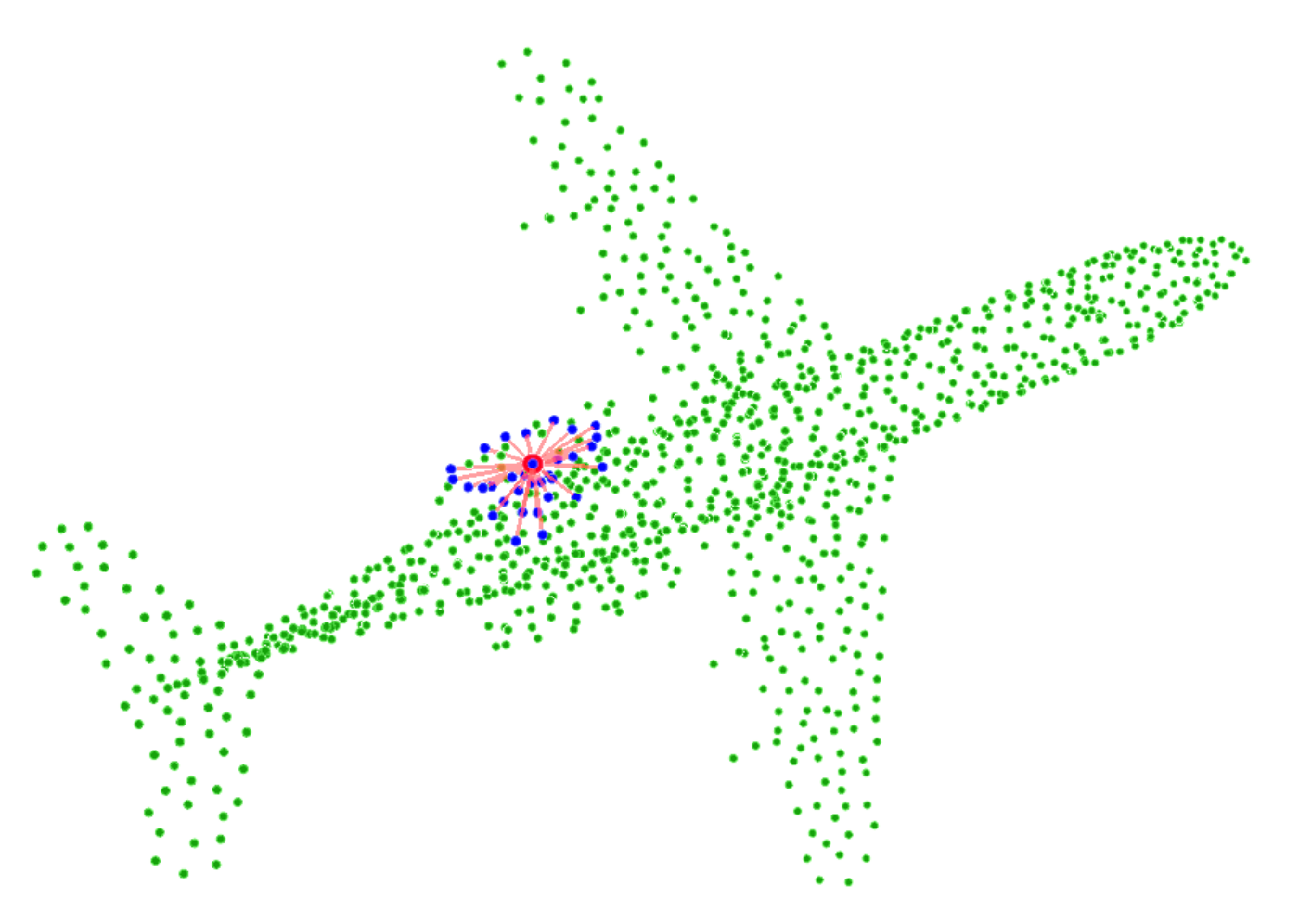}
\end{center}
  \caption{Airplane: \textcolor{green}{Green} points - input point cloud, \textcolor{red}{Red} point - anchor point for a neighborhood obtained from the farthest point sampling step, \textcolor{blue}{Blue} points - points queried from the neighborhood of anchor point, \textcolor{red}{Red} Vectors - Directional Vectors from the anchor point to the neighbor points. The magnitude of each directional vector represents the distance of the neighbor point to its anchor point. The direction of these vectors is from the anchor point to the neighbor point.}
\label{Fig1}
\end{figure}

\section{Method}
In this section, we elaborate on the usage of distances and directional vectors from the anchor point to neighborhood points already computed in the grouping stage of classification models from PointNet++~\cite{qi2017pointnet++} to the most recent PointNeXt~\cite{PointNeXt} method.

\subsection{Neighborhood point distance}
With the inception of PointNet++~\cite{qi2017pointnet++}, which introduced neighborhood querying using a fixed radius ball (as shown in Fig.~\ref{Fig1}) or k-nearest neighbor(kNN), several methods have continued to take advantage of these querying techniques. In the case of ball-queried neighborhood points, the distance between the anchor and neighborhood points is computed to determine if the given point is within the ball of radius $r$. i.e., a point $(x_{i}, y_{i}, z_{i})$  is selected from the input point set $xyz_1$, if it lies within the ball of radius $r$ centered at ($x_{j}$, $y_{j}$, $z_{j}$) (anchor points are obtained as $xyz_2$ using the farthest point sampling method on the input $xyz_1$ set). Euclidean distance $d_{i}$ of the point from the anchor point/center of the ball is calculated and the distance is normalized with the radius $r$ as shown in Eq.~\ref{first_equation}. When $d_{i}$ is less than 1, the point is within the ball; if the distance is equal to 1, then the point is on the ball. And if the distance is greater than 1, the point is outside the ball. A point $(x_{i}, y_{i}, z_{i})$ with distance $d_{i}$ less than 1 is selected. We save this distance and use it as an additional neighborhood feature.

\begin{equation}\label{first_equation}
d_{i}\; {=} \;
\frac{\sqrt{(x_{i} - x_{j})^2 + (y_{i} - y_{j})^2 + (z_{i}-z_{j})^2} }{r}\;
\end{equation}
\vspace{-1em}
\subsection{Directional Vectors as neighborhood features}

\begin{figure}[htbp]
\centering
\begin{tabular}{cc}
\includegraphics[scale=0.22]{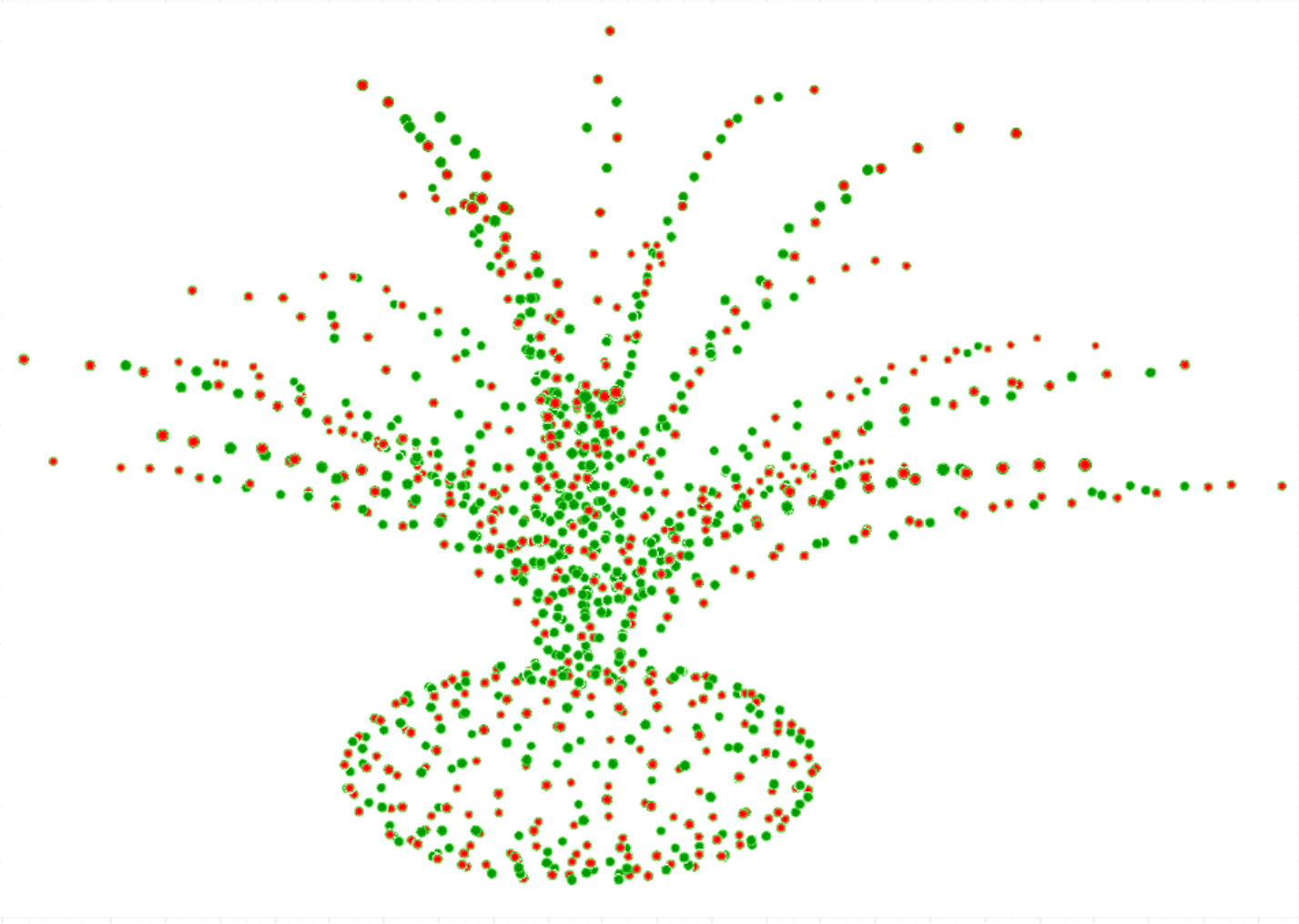} %0.418
&
\includegraphics[scale=0.22]{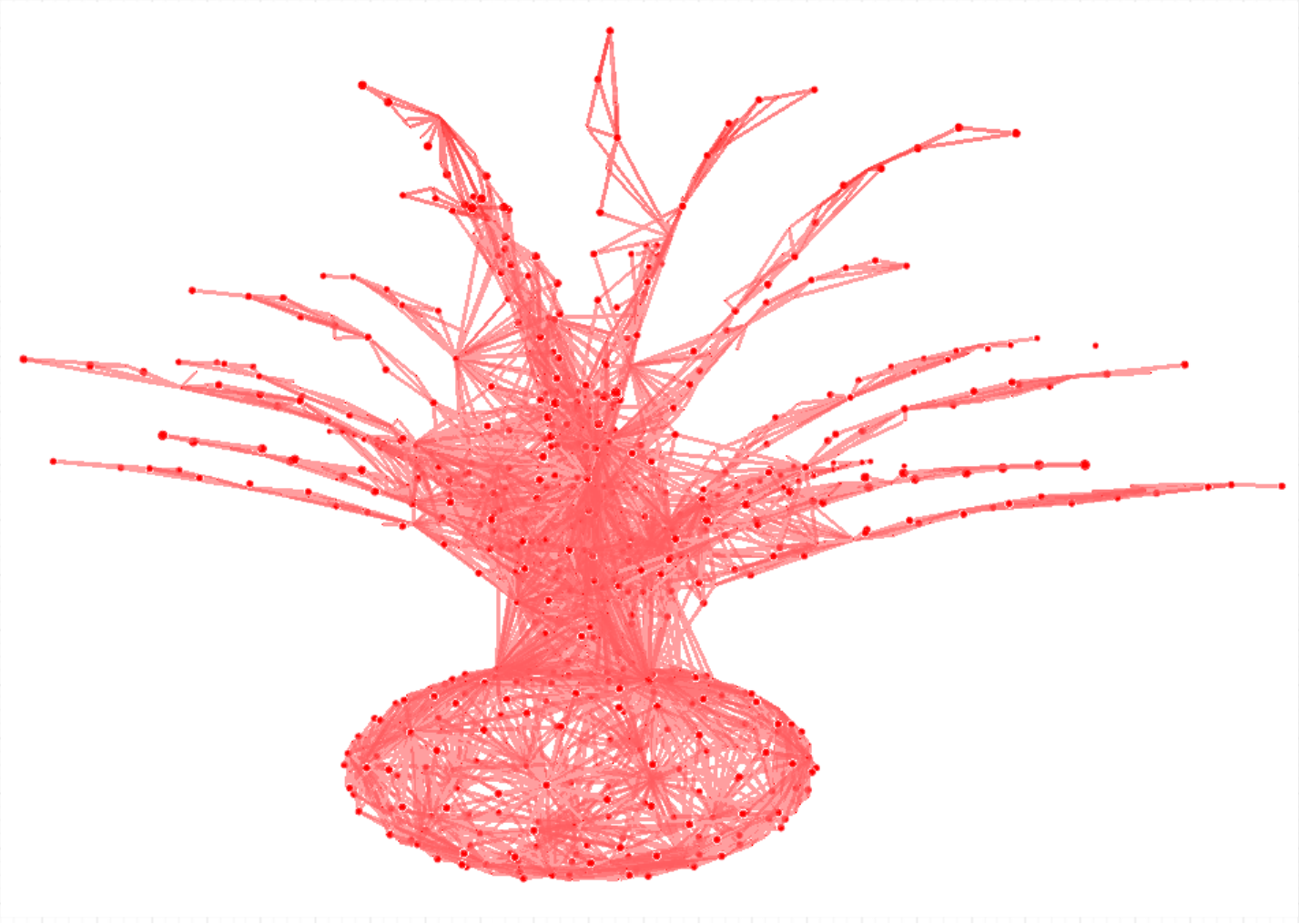} %0.418
\end{tabular}
\caption{Left: Input point cloud (\textcolor{green}{green} points), Farthest Point Sampled points (\textcolor{red}{red} points), Right: Directional vectors at each of the \textcolor{red}{red} points.}
\label{Fig2}
\end{figure}

\begin{figure}[htbp]
\begin{center}
%\fbox{\rule{0pt}{2in} \rule{0.9\linewidth}{0pt}}
\includegraphics[width=0.9\linewidth]{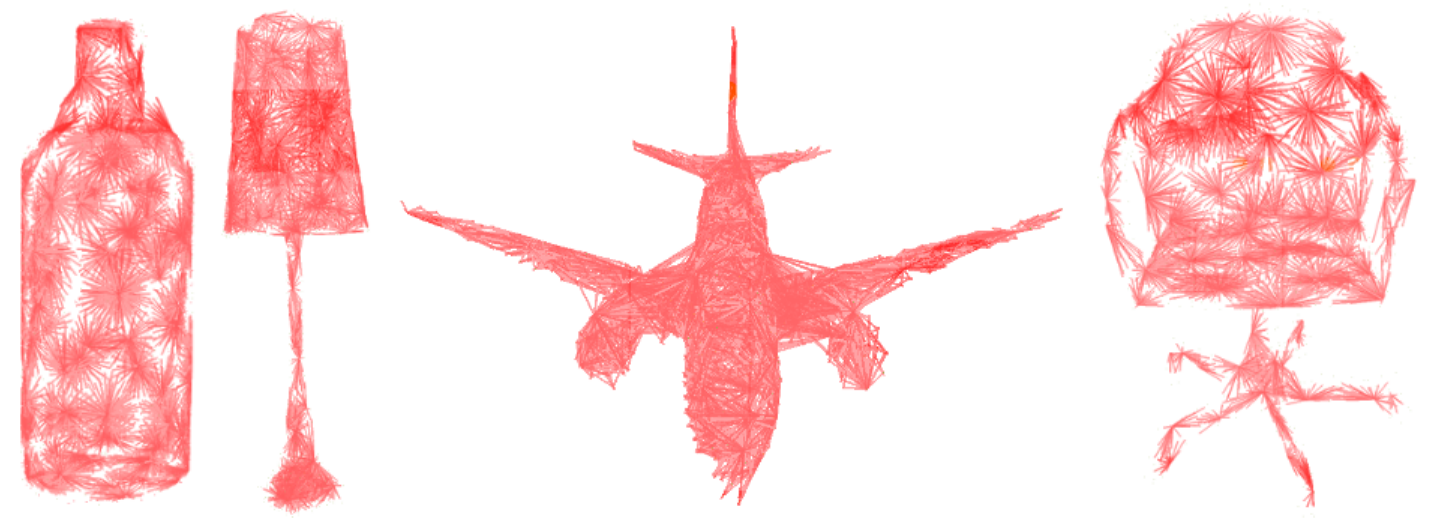}
\end{center}
  \caption{Few more visual examples of 3D objects with directional vectors after the grouping stage. From Left to Right: bottle, lamp, airplane, and chair.}
\label{fig3}
\end{figure}

\vspace{-1em}
After querying, the neighborhood points are centered at the anchor point by subtracting the anchor point from the neighborhood points. By doing so, the directional vectors ($dv_j$) at the grouping level are computed in models that adopted the grouping layer of PointNet++~\cite{qi2017pointnet++} (check Fig~\ref{Fig2}b). PointNeXt~\cite{PointNeXt} normalizes directional vectors using the radius $r$ as shown in Eq.~\ref{second_equation}. These radius normalized-directional vectors are mapped to relatively higher dimensions but are not directly included as features while hierarchically learning the discriminative features. Instead, we include these directional vectors as neighborhood-level features by concatenating these vectors to the relatively higher dimensional features in a given neighborhood. Figure~\ref{fig3} shows the visualization of directional vectors.

\begin{equation}\label{second_equation}
dv_j = \frac{[(x_{i} - x_{j}), (y_{i} - y_{j}), (z_{i}-z_{j})]}{r}\;
\end{equation}

\vspace{-1em}

\subsection{Weights-averaging of checkpoints}
Several methods save the model during training at every increment in the overall or mean accuracy and only use the best checkpoint for inference. We observe the best results of the PointNeXt model on the ScanObjectNN dataset around 190 to 210 epochs in a training cycle of 250. We noticed several epochs past the 210th epoch, achieved comparable results to the best evaluation at the 210th epoch. Following the idea of weights averaging from model soups~\cite{wortsman2022model}, we take the best checkpoints from the training session, average the weights, and use the weight-averaged checkpoint for inference. Weight averaging benefited in stabilizing the weights and greatly improved the test accuracy. 

\vspace{-1em}
% \newpage
\section{Experiments}
We evaluate the impact of additional local neighborhood features i.e., radius normalized-distance and directional vectors, in the PointNeXt model using the benchmark synthetic dataset ModelNet40~\cite{wu20153d}. ModelNet40 dataset has 12,311 shapes spread across 40 classes. We use the dataset's official split of 9,843 objects for training and 2,468 for testing. Additionally, we evaluate the PointNeXt model using two robust real-world datasets, i.e., the ScanObjectNN~\cite{uy2019revisiting} and 3DGrocery100~\cite{sheshappanavar2024benchmark} (For more details refer the project page:~\href{https://bigdatavision.org/3DGrocery100/}{https://bigdatavision.org/3DGrocery100/}). 

ScanObjectNN contains 14,510 objects obtained by perturbing the original 2,902 objects. ScanObjectNN is derived from 700 unique scenes from two popular scene meshes datasets; SceneNN~\cite{hua2016scenenn} with 100 objects and ScanNet~\cite{dai2017scannet} with 1,513 objects. Although the ScanObjectNN dataset has six variants, we use the hardest variant represented by PB\_T50\_RS (following PointNeXT and other recent models).  PB\_T50\_RS variant constitutes perturbed objects with 50\% translation and involves rotation and scaling. 

3DGrocery100 dataset at a very high level contains three categories, i.e., Fruits, Vegetables, and Packages with 34, 28, and 38 classes, 37,587, 27,707, and 22,604 point clouds, respectively. The Fruits subset named Apple10 contains ten apple classes and 24 non-apple fruit classes. We consider each of the three categories, i.e., Fruits (non-apples), Vegetables, and Packages, along with the Full dataset as subset datasets. We train the PointNeXT model with and without our proposed additional local neighborhood features using the five subset datasets of 3DGrocery100~\cite{sheshappanavar2024benchmark}. We evaluate the impact of the local neighborhood features in improving the classification results on these subsets of the grocery dataset on the PointNeXt model. Finally, we discuss and analyze our experimental results on Apple10, Fruits, Vegetables, Packages, and Full subsets. 

\section{Experimental Results}
In this section, we show the benefits of using the radius $r$-normalized distance and directional vectors as additional neighborhood features in the PointNeXt model when trained and tested on ScanObjectNN, ModelNet40, and 3DGrocery100 datasets. Our experimental results encourage using these already computed features in future models.

% \subsection{Classification of ScanObjectNN Dataset}

Table~\ref{tab1} shows the impact of radius normalized distance and directional vector concatenation to local neighborhood features in the PointNeXt model trained on the ScanObjectNN dataset. In addition, we show inference results by averaging the weights of the best two checkpoints. Using the PointNeXt~\cite{PointNeXt} settings, we train PointNet++ without and with the additional local neighborhood features and report 3D classification accuracy. 

\begin{table}
\centering
\caption{Additive study of sequentially adding our strategies for classification on ScanObjectNN.}\label{tab1}
\begin{tabular}{l|c|c|c|c}
\hline
Improvements &  Overall Accuracy (\%)& $\Delta$ & Mean Average Accuracy (\%)&$\Delta$\\
\hline
PointNeXt~\cite{PointNeXt} & 87.7 $\pm$ 0.4 & - &85.8 $\pm$ 0.6& - \\
+ $r$-normalized distance & 87.9 $\pm$ 0.4 & +0.2& 86.4 $\pm$ 0.3& +0.6\\
+ directional vectors & 88.2 $\pm$ 0.2 & +0.3& 86.9 $\pm$ 0.2& +0.5 \\
+ best two average & 88.4 $\pm$ 0.2 & +0.2&87.1 $\pm$ 0.3& +0.2 \\ 
\hline
Overall Best & 88.6 & \textbf{0.5} $\uparrow$ & 87.4 & \textbf{1.0} $\uparrow$\\
\hline
PointNet++&83.5&-&81.7&-\\
PointNet++ (+ours)&86.0&\textbf{+2.5}$\uparrow$&84.1&\textbf{+2.4}$\uparrow$\\
\hline
Pix4Point~\cite{qian2022pix4point} & 86.2 & - & 83.9 & -\\
Pix4Point~\cite{qian2022pix4point}(+ours) & 86.4 & \textbf{+0.2}$\uparrow$ & 84.0 & \textbf{+0.1}$\uparrow$ \\
\hline
\end{tabular}
\end{table}

% \subsection{Classification of ModelNet40 Dataset}

\begin{table}
\centering
\caption{ 3D Classification accuracy of PointNeXt model with our approach on ModelNet40. OA = Overall Accuracy, mAcc = Mean Average Accuracy}\label{tab2}
\begin{tabular}{l|c|c|c|c}
\hline
Improvements &  OA (\%)& $\Delta$ & mAcc (\%)&$\Delta$\\
\hline
PointNet++ & 90.7 & - & 86.6 & -\\
PointNet++(+ours) & 91.1 & \textbf{+0.4} & 89.1 & \textbf{+2.5}\\
\hline
PointNeXt~\cite{PointNeXt} & 93.3 & - &91.0& - \\
PointNeXt + (ours) & 93.5 &\textbf{+0.2}& 91.0&\textbf{+0.0} \\
% Overall Best & &  & & \\
\hline
\end{tabular}
\end{table}

% \subsection{Classification of 3D Grocery Dataset}
Table~\ref{tab2} shows a comparable overall improvement in the classification accuracy of the PointNeXt model trained on the ModelNet40 dataset. Most recent models are saturated with ModelNet40 results (in the range of 93-94), and an improvement of 0.2\% is good. Table~\ref{tab3} shows the impact of radius normalized distance and directional vector concatenation to neighborhood features in the PointNeXt model trained on all five subsets of the 3DGroceyr100 dataset~\cite{sheshappanavar2024benchmark} (without colors). Improvements in classification accuracy on synthetic and real-world datasets show the broad applicability of the additional features and the novelty of our work.

\begin{table}[!h]\centering
  \caption{3D point cloud classification on PointNeXt model on all subsets of 3DGrocery100~\cite{sheshappanavar2024benchmark}.}
  \begin{adjustbox}{width=0.7\linewidth}
  \begin{tabular}{c|c|c|c|c|c}
    \toprule
    \textbf{Models$\downarrow$/Subsets$\rightarrow$}&\textbf{Apple10}&\textbf{Fruits}&\textbf{Vegetables}&\textbf{Packages}&\textbf{Full}\\
    \midrule
    \#Classes &10&24&28&38&100\\
    \hline 
    % \#Images &1025&2586&3029&4115&10755\\
    % \midrule
    \#Point Clouds &12905&24682&27707&22604&87898\\
    \hline
    Train&9706&18406&20720&17214&66032\\
    Test&3199&6276&6987&5390&21866\\
    \hline
PointNeXt~\cite{PointNeXt}&21.6&40.6&48.4&81.4&47.7\\
PointNeXt+(ours)&22.6&45.4&51.8&83.0&50.5\\
\midrule
$\uparrow$&\textbf{+1.0}&\textbf{+4.8}&\textbf{+3.4}&\textbf{+1.6}&\textbf{+2.8}\\
    \hline
  \end{tabular}
  \end{adjustbox}
  \label{tab3}
\end{table}

\section{Ablation Study}

\subsection{Distance vs. $r$-normalized distance}
From Equation~\ref{first_equation}, we know the distance of the neighborhood points to the anchor points is normalized using the radius $r$ used for querying the points. While existing methods only compute the distance to check whether a point is within the ball for querying, here we provide experimental results to show that the normalized distance as an additional neighborhood feature gives better mean accuracy, as shown in Table~\ref{tab4}.
% \vspace{-1em}
\begin{table}[!h]
\centering
\caption{Comparison of distance vs. radius $r$-normalized distance as an additional neighborhood feature. Model Used: PointNeXt~\cite{PointNeXt} and Dataset Used: ScanObjectNN's hardest variant.}\label{tab4}
\begin{tabular}{c|c|c}
\hline
Local Features &  OA (\%) & mAcc (\%) \\
\hline
distance & 88.3 & 86.5\\
$r$-normalized distance & 88.3 & 86.7\\
\hline
\end{tabular}
\end{table}

\subsection{Weight averaging (same training session)}
During a training session, we saved the checkpoints at the top 15 best results. A systematic study of averaging the weights of these checkpoints during inference is shown in Table~\ref{tab5}. From our study, we gathered that averaging the weights of the top two checkpoints further improved classification accuracy. Unlike the ensemble of models, averaging weights and saving the averaged model does not consume extra storage space.

% \vspace{-1em}
\begin{table}[!h]
\centering
\caption{ Ablation study on weight averaging of checkpoints for classification on ScanObjectNN.}\label{tab5}
\begin{tabular}{l|c|c|c|c}
\hline
Improvements &  Overall Accuracy (\%)& $\Delta$ & Mean Average Accuracy (\%)&$\Delta$\\
\hline
PointNeXt & 88.1 & - &86.4& - \\
\hline
\multicolumn{5}{c}{Weight averaging (of checkpoints) from same training session}\\
\hline
+ top-2 & 88.4&\textbf{+0.3}&87.1&\textbf{+0.7} \\
+ top-3 & 88.3&+0.2&87.1&+0.7\\
+ top-5& 88.2&+0.1&86.3&-0.1\\ 
+ top-10& 88.0&-0.1&86.4&+0.0\\ 
+ top-15& 87.9&-0.2&86.3&-0.1\\ 
\hline
\end{tabular}
\end{table}
\vspace{-2em}

\begin{table}
\centering
\caption{Cost of additional neighborhood features in terms of the number of parameters, GFLOPs, and Throughput. Tested with 1024 input points and a batch size of 128.}\label{tab6}
\begin{tabular}{l|c|c|c}
\hline
 &  PointNeXt~\cite{PointNeXt}& PointNeXt + (ours)&$\Delta$\\
\hline
Params (M)&1.3671&1.3690&+0.0019\\
GFLOPs&1.64&1.65&+0.01\\
Throughput (ms/sample)&0.042&0.045&+0.003\\
\hline
\end{tabular}
\end{table}
\vspace{-2em}
\subsection{Cost of additional neighborhood features}
The cost of adding local neighborhood features is negligible based on three measures: the number of parameters (in millions), GFLOPs, and throughput (millisecond/sample). Despite the four additional features - radius $r$-normalized distance (1) and directional vectors (3) - already computed in the set abstraction layers, there is an increase in all three measures, as shown in Table~\ref{tab6}. The throughput in milliseconds (ms) per sample increases by a negligible 0.003 (i.e., 3 microseconds/sample) due to the division operation (to normalize the distance) and additional network computations from the extra features. The overall computational cost is still below the PointNet++ model (1.69 GFLOPs as shown in table 6 of TR-Net~\cite{liu2022tr}). Improvements in the accuracy and a minimal increase in the number of parameters, GFLOPs, and computational costs are encouraging.

%------------------------------------------

\section{Conclusion}
In this paper, we demonstrate the use of local neighborhood features as additional local features for point cloud classification. We also introduce an inference strategy to benefit the overall results. We present the benefits of neighborhood features and inference strategy on the state-of-the-art PointNeXt model using three datasets: ModelNet40, ScanObjectNN, and 3DGrocery100. The neighborhood features significantly improved the classification accuracy, particularly in the case of real-world datasets. Improvements of \textbf{0.5\%}, \textbf{1\%}, \textbf{4.8\%}, \textbf{3.4\%}, \textbf{1.6\%}, \textbf{2.8\%}, and \textbf{0.2\%} in the overall accuracy on the real-world ScanObjectNN (hardest variant), 3DGrocery100 dataset's subsets Apple10, Fruits, Vegetables, Packages, Full, and synthetic dataset ModelNet40, respectively on PointNeXt model are significant. Our ablation study provides a detailed analysis to help understand the trade-offs in using additional local neighborhood features. Our results and ablation study strongly encourage the use of these features in future models.

\end{document}